\pdfoutput=1

\documentclass[11pt]{article}

\usepackage[]{acl}

\usepackage{times}
\usepackage{latexsym}

\usepackage[T1]{fontenc}

\usepackage[utf8]{inputenc}

\usepackage{microtype}
\usepackage{graphicx}
\usepackage{hyperref}

\title{Random Text Perturbations Work, but not Always}

\author{Zhengxiang Wang \\ Department of Linguistics, Stony Brook University \\
  \texttt{zhengxiang.wang@stonybrook.edu}}

\begin{document}
\maketitle
\thispagestyle{plain}
\pagestyle{plain}
\begin{abstract}

We present three large-scale experiments on binary text matching classification task both in Chinese and English to evaluate the effectiveness and generalizability of random text perturbations as a data augmentation approach for NLP. It is found that the augmentation can bring both negative and positive effects to the test set performance of three neural classification models, depending on whether the models train on enough original training examples. This remains true no matter whether five random text editing operations, used to augment text, are applied together or separately. Our study demonstrates with strong implication that the effectiveness of random text perturbations is task specific and not generally positive. 

\end{abstract}

\section{Introduction}

Data augmentation (DA) is a common strategy to generate novel label-preserving data to remedy data scarcity and imbalance problems \cite{xie2020unsupervised}, which has been applied with noteworthy success in image and speech recognition \cite{Iwana2021,Park2019,Shorten2019ASO}. In the field of natural language processing (NLP), there have also been a number of studies that use various DA techniques to boost the trained models’ performance \cite{feng-etal-2021-survey,Liu:9240734}, ranging from word replacement \cite{wang-yang-2015-thats,wang-etal-2018-switchout,Zhang2015}, to predictive neural language models \cite{hou2018,kobayashi-2018-contextual,Kurata2016LabeledDG}. However, an evident and critical difference between text and image/speech is that text cannot be treated as purely physical. For any given sequence of words, both the word order and the semantic compatibility among words affect the meaning, and possibly the label of the sequence. This complex nature raises the question as to whether there exists some generally effective DA approach for NLP because automatic strict paraphrasing barely exists \cite{bhagat-hovy-2013-squibs}. 

\begin{table}[t]
\footnotesize
\centering
\flushleft
\begin{tabular}{l|p{5.4cm}}
\hline
\textbf{Operation} & \textbf{Text} \\ \hline
None & A sad, superior human comedy played out on the back roads of life.\\ \hline
SR & A sad, superior \textbf{\textit{homo}} \textbf{\textit{funniness}} played out on the back roads of life. \\ \hline
RI &  A sad, superior human comedy played \textbf{\textit{man}} out on the back \textbf{\textit{stunned}} roads of life. \\ \hline
RS & \textbf{\textit{the}} sad, superior human comedy played out on \textbf{\textit{roads}} \textbf{\textit{back}} \textbf{\textit{A}} of life. \\ \hline
RD & A superior human comedy played out on back roads life. \\ \hline
\end{tabular}
\caption{\label{tab:1}
Text augmented with two edits each DA technique by EDA. The original text is from \citet{wei-zou-2019-eda}. SR: Synonym Replacement; RI: Random Insertion; RS: Random Swap; RD: Random Deletion.}
\end{table}

This study is a preliminary examination of the effectiveness and generalizability of random text perturbations as a DA approach, exemplified by Easy Data Augmentation (EDA)\footnote{\url{https://github.com/jasonwei20/eda_nlp}}, which has been proposed to be a universal DA approach for NLP \cite{wei-zou-2019-eda}. This approach consists of four commonly used token-level editing operations \cite{wei-etal-2021-text,wei-zou-2019-eda}, i.e., Synonym Replacement (SR), Random Insertion (RI), Random Swap (RS), and Random Deletion (RD). SR randomly replaces synonyms for eligible words, while RS randomly swap word pairs. RI inserts random synonyms, if any, instead of random words, whereas RD deletes words at random. Simple as these operations may seem, they have shown general success in various sentiment-related and sentence type classification tasks \cite{wei-zou-2019-eda}.

To do the examination, we first present a linguistically informed hypothesis and propose a relevant method of evaluation in section~\ref{sec:2}. We then introduce the experimental settings and results in section~\ref{sec:3} and section~\ref{sec:4}, respectively. The paper ends with some discussions and conclusions in section~\ref{sec:5}. 

The major contributions of this study are threefold. First, it reveals the possible inherent limitations of random text perturbations used as a DA approach for NLP with cross-lingual evidence. Second, the paper provides a critical angle and possibly a general way to evaluate the effectiveness and generalizability of a DA approach or technique for NLP. Third, we present an EDA-like Python program that refines EDA's functionalities, contains a novel DA technique, and can be easily employed for text augmentation in other languages. The source code for this program can be found at \url{https://github.com/jaaack-wang/reda}.

\section{\label{sec:2}Hypothesis and evaluation method}

From a linguistic point of view, the success of EDA defies understanding, as the augmented texts produced by EDA can often be unnatural, ungrammatical, or meaningless, such as examples shown in Table~\ref{tab:1}. However, it is also not surprising that these imperfect augmented texts may help models generalize better on test sets for some simple text classification tasks, as they introduce certain noise to the training examples that reduces overfitting while not damaging key information, which can easily lead to label change. For example, for sentence-level sentiment analysis, the sentiment of a sentence is often captured by only few keywords \cite{Bing2012}. It follows, as long as an augmented text keeps these few keywords or similar replaced words, it still reasonably preserves the sentiment label of the original text even if it is a problematic sentence. That explains the decline in models’ performance in the ablation experiments by \citet{wei-zou-2019-eda}, where SR and RD were applied with 30\% or larger editing rate, making the key lexical features more likely to be replaced or deleted. In contrast, RS and RI were overall harmless no matter how large proportion of a text was edited. This is simply because unlike SR and RD, RS and RI do not remove any lexical items in the original texts. 

Therefore, we hypothesize that the effectiveness of random text perturbations is task specific and thus may not constitute a generally effective DA approach for NLP, especially if the task requires stricter semantic equivalence of the augmented text to the original text. To verify this hypothesis, we conduct experiments on binary text matching classification task both in Chinese and in English to see if five simple text editing operations, adapted from EDA, can improve the performance of three commonly used deep learning models. Since text matching classification involves prediction of whether a text pair match in meaning, it is inherently a more reliable way to test if a certain level of semantic changes, caused by text perturbations, can remain useful for training NLP models. 

\section{\label{sec:3}Experimental settings}

\subsection{Datasets}

We used two large-scale benchmark datasets, the Large-scale Chinese Question Matching Corpus (LCQMC) compiled by \citet{liu-etal-2018-lcqmc} and the Quora Question Pairs Dataset (QQQD)\footnote{\url{https://quoradata.quora.com/First-Quora-Dataset-Release-Question-Pairs}}, to represent binary text matching task in Chinese and in English, respectively. Both datasets contain a large collection of question pairs manually annotated with a label, 0 or 1, to indicate whether a pair match or not in terms of the expressed intents. 

For LCQMC, we reused the original train, development, and test sets as provided by the authors \cite{liu-etal-2018-lcqmc}. For QQQD, we created three label-balanced data sets based on its train set since the test set is made unlabeled for online competition. The basic statistics about these two datasets are given in Table~\ref{tab:2}.

\begin{table}
\footnotesize
\centering
\begin{tabular}{lcc}
\hline
\textbf{Split} & \textbf{LCQMC} & \textbf{QQQD} \\
 & \scriptsize (Matched \& Mismatched) & \scriptsize (Matched \& Mismatched) \\ \hline
Train & 238,766 & 260,000 \\
 & \scriptsize (138,574 \& 100,192)  & \scriptsize (130,000 \& 130,000) \\ \hline
Dev & 8,802 & 20,000 \\
 &  \scriptsize (4,402 \& 4,400) & \scriptsize (10,000 \& 10,000)  \\ \hline
Test & 12,500 & 18,526 \\
 & \scriptsize (6,250 \& 6,250) & \scriptsize (9,263 \& 9,263)  \\ \hline
 \end{tabular}
\caption{\label{tab:2}
Statistics of the LCQMAC \& QQQD data sets.}
\end{table}

\subsection{Augmentation Setup}

We created REDA (i.e., Revised EDA), a Python program adapted from EDA, to perform text augmentation in this study. REDA comes with the four text editing operations as in EDA, but also presents a novel technique called Random Mix (RM), which randomly selects 2-4 of the other four operations to further diversify the augmented texts. Besides, the rationales for REDA over EDA are as follows: unlike EDA, (1) REDA has a mechanism to prevent deduplicates, which can occur when there are no synonyms to replace (SR) or insert (RS) for words in the original text, or when the same words are replaced or swapped back during SR and RS operations. (2) REDA does not preprocess the input text (e.g., removing punctuations and stop words), which we believe are more in line with the basic idea of random text perturbations, the focus of this study. (3) REDA only replaces one word with its synonym at a given position at a time, instead of all its occurrences, which we see as extra edits. (4) REDA supports Chinese text augmentation in addition to English text augmentation.

Due to costs of doing experiments at this scale, we are unable to evaluate the effects of different initializations of REDA (e.g., editing rate) on the trained models’ performance. Therefore, we initialized REDA with small editing rates, among others, based on our hypothesis and \citet{wei-zou-2019-eda}, which we believe is reasonably informed to reveal the effectiveness of random text perturbations for our experiments in general. Please refer to Appendix \hyperref[sec:A]{A} for details. 

\begin{table}
\footnotesize
\centering
\begin{tabular}{lccccc}
\hline
\textbf{Model} & \textbf{5k} & \textbf{10k} & \textbf{50k} & \textbf{100k} & \textbf{Full set} \\ \hline
CBOW & 59.4\% & 60.4\% & 65.4\% & 67.8\% & 73.8\% \\
\hspace{0.1cm} \scriptsize + REDA & 58.1\%  & 60.9\% & \textbf{68.2\%} & \textbf{72.2\%} & \textbf{76.4\%} \\ \hline

CNN & 59.3\% & \textbf{63.4\%} & 67.2\% & 69.0\% & 72.9\% \\
\hspace{0.1cm} \scriptsize + REDA & 59.8\% & 62.6\% & 66.8\% & 69.8\% & 74.9\% \\ \hline

LSTM & \textbf{60.0\%} & 62.1\% & 66.2\% & 69.6\% & 74.8\% \\
\hspace{0.1cm} \scriptsize + REDA & 58.9\% & 61.5\% & 67.7\% & 71.8\% & \textbf{76.4\%} \\ \hline

Average & 59.6\% & 62.0\% & 66.3\% & 68.8\% & 73.8\% \\
\hspace{0.1cm} \scriptsize + REDA & 58.9\% & 61.7\% & 67.6\% & 71.3\% & 75.9\% \\ \hline

\end{tabular}
\caption{\label{tab:3}
Test set accuracy scores of the three models trained on LCQMC’s train sets of varying size with and without augmentation.}
\end{table}
\begin{table}
\footnotesize
\centering
\begin{tabular}{lccccc}
\hline
\textbf{Metric} & \textbf{5k} & \textbf{10k} & \textbf{50k} & \textbf{100k} & \textbf{Full set} \\ \hline
Precision & 57.2\% & 59.2\% & 62.4\% & 64.1\% & 68.2\% \\
\hspace{0.1cm} \scriptsize + REDA & 56.9\%  & 59.7\% & 63.9\% & 66.5\% & 70.2\% \\ \hline

Recall & 75.5\% & 77.3\% & 82.0\% & 85.5\% & 89.2\% \\
\hspace{0.1cm} \scriptsize + REDA & 73.6\% & 72.1\% & 80.7\% & 85.5\% & 90.0\% \\ \hline

\end{tabular}
\caption{\label{tab:4}
Average test set precision and recall scores of the three models trained on LCQMC’s train sets of varying size with and without augmentation.}
\end{table}

\subsection{Classification Models}

We chose three common neural models, including Continuous Bag of Word (CBOW) model, Convolutional Neural Network (CNN) model, and Long Short-Term Memory (LSTM) model, as the classification models. The models were trained with a 64 batch size, a fixed .0005 learning rate, and constantly 3 epochs. We used Adaptive Moment Estimation (Adam) as the optimizer and cross entropy as the loss function. Also, unlike \citet{wei-zou-2019-eda}, we did not utilize pretrained word embeddings for our models, which will make the effects of text perturbations complicated and less interpretable. Plus, we believe for a DA approach to be generally effective, it should also work in a setting where resources for pretrained word embeddings are limited or unavailable. 

The details of the model configurations and the training settings are provided in Appendix \hyperref[sec:B]{B}.

\begin{table}
\footnotesize
\centering
\begin{tabular}{lccccc}
\hline
\textbf{Model} & \textbf{10k} & \textbf{50k} & \textbf{100k} & \textbf{150k} & \textbf{Full set} \\ \hline
CBOW & 64.4\% & 69.9\% & 72.1\% & 74.2\% & 77.7\% \\
\hspace{0.1cm} \scriptsize + REDA & 62.5\%  & 68.5\% & 71.6\% & 74.8\% & 78.0\% \\ \hline

CNN & \textbf{66.1\%} & 71.1\% & 72.6\% & 73.4\% & 75.9\% \\
\hspace{0.1cm} \scriptsize + REDA & 63.7\% & 69.9\% & 72.7\% & \textbf{75.3\%} & 77.6\% \\ \hline

LSTM & 65.7\% & \textbf{71.6\%} & \textbf{72.9\%} & 75.0\% & 77.9\% \\
\hspace{0.1cm} \scriptsize + REDA & 64.0\% & 69.8\% & 72.5\% & 75.1\% & \textbf{78.1\%} \\ \hline

Average & 65.4\% & 70.9\% & 72.5\% & 74.2\% & 77.2\% \\
\hspace{0.1cm} \scriptsize + REDA & 63.4\% & 69.4\% & 72.3\% & 75.1\% & 77.9\% \\ \hline

\end{tabular}
\caption{\label{tab:5}
Test set accuracy scores of the three models trained on QQQD’s train sets of varying size with and without augmentation.}
\end{table}
\begin{table}
\footnotesize
\centering
\begin{tabular}{lccccc}
\hline
\textbf{Metric} & \textbf{10k} & \textbf{50k} & \textbf{100k} & \textbf{150k} & \textbf{Full set} \\ \hline
Precision & 63.8\% & 70.2\% & 71.1\% & 72.4\% & 75.6\% \\
\hspace{0.1cm} \scriptsize + REDA & 61.8\%  & 67.6\% & 70.5\% & 74.2\% & 76.4\% \\ \hline

Recall & 71.4\% & 72.5\% & 76.1\% & 78.3\% & 80.2\% \\
\hspace{0.1cm} \scriptsize + REDA & 70.4\% & 74.3\% & 76.7\% & 76.9\% & 80.9\% \\ \hline

\end{tabular}
\caption{\label{tab:6}
Average test set precision and recall scores of the three models trained on QQQD’s train sets of varying size with and without augmentation.}
\end{table}

\section{\label{sec:4}Results}

This section reports the test set performance of the three classification models trained on train sets of varying size with and without augmentation for the binary text matching task in Chinese and in English. We used accuracy as the main metric to evaluate the effectiveness of random text perturbations. The average precision and recall scores of the three models are taken as secondary metrics for more nuanced analyses. Due to the experimental costs, we only did ablation study on LCQMC to examine the effectiveness of the five DA techniques applied separately. The classification results on the original train sets are seen as baselines. Please refer to Appendix \hyperref[sec:C]{C} for the size of augmented train sets.

\subsection{For Chinese}

As can be seen in Table~\ref{tab:3}, the size of the train set affects whether models trained on the augmented train sets outperform the baselines, with the threshold being near 50k (about 21\% of the original full train set). Table~\ref{tab:4} shows that the gains in the test set accuracy scores are mainly driven by two factors: (1) the leading precision scores of the REDA-led models after the 10k training size; (2) the narrowing gap in the recall scores after the 50k training size. That implies, the classification models learn to make less false positives with sufficient original training examples augmented. But before the threshold, augmentation is nevertheless detrimental to the models’ performance even with the drastic increase of the training examples.

\subsection{For English}

Table~\ref{tab:5} resembles Table~\ref{tab:4} in data patterns, reaffirming the need of sufficient training examples for random text perturbations to work for the binary text matching task. The threshold, however, is much larger this time, nearing the 150k training size (about 57\% of the original full train set), which may be dataset specific. Moreover, the REDA-led models only outperform the baselines by a small margin on average (i.e., less than 1\%) on the test set, smaller than the previous section. Table~\ref{tab:6} also shows that the increasing test set precision and recall scores, particularly the former, account for the performance gains of the REDA-led models.

\subsection{Ablation Study: each DA technique}

With random text perturbations requiring ample original training examples to be effective as presented above, a natural question becomes: what if the five DA techniques were applied separately? To get a more nuanced and reliable observation, we augmented train sets of 11 different sizes, instead of 5 as in the previous sections. These 11 training sizes roughly correspond to 2\%, 4\%, 10\%, 21\%, 31\%, 42\%, 52\%, 63\%, 73\%, 84\%, and 100\% of the LCQMC’s train set, respectively.

\begin{figure}[!htb]
\centering
\captionsetup{font=small}
  \includegraphics[width=1\columnwidth]{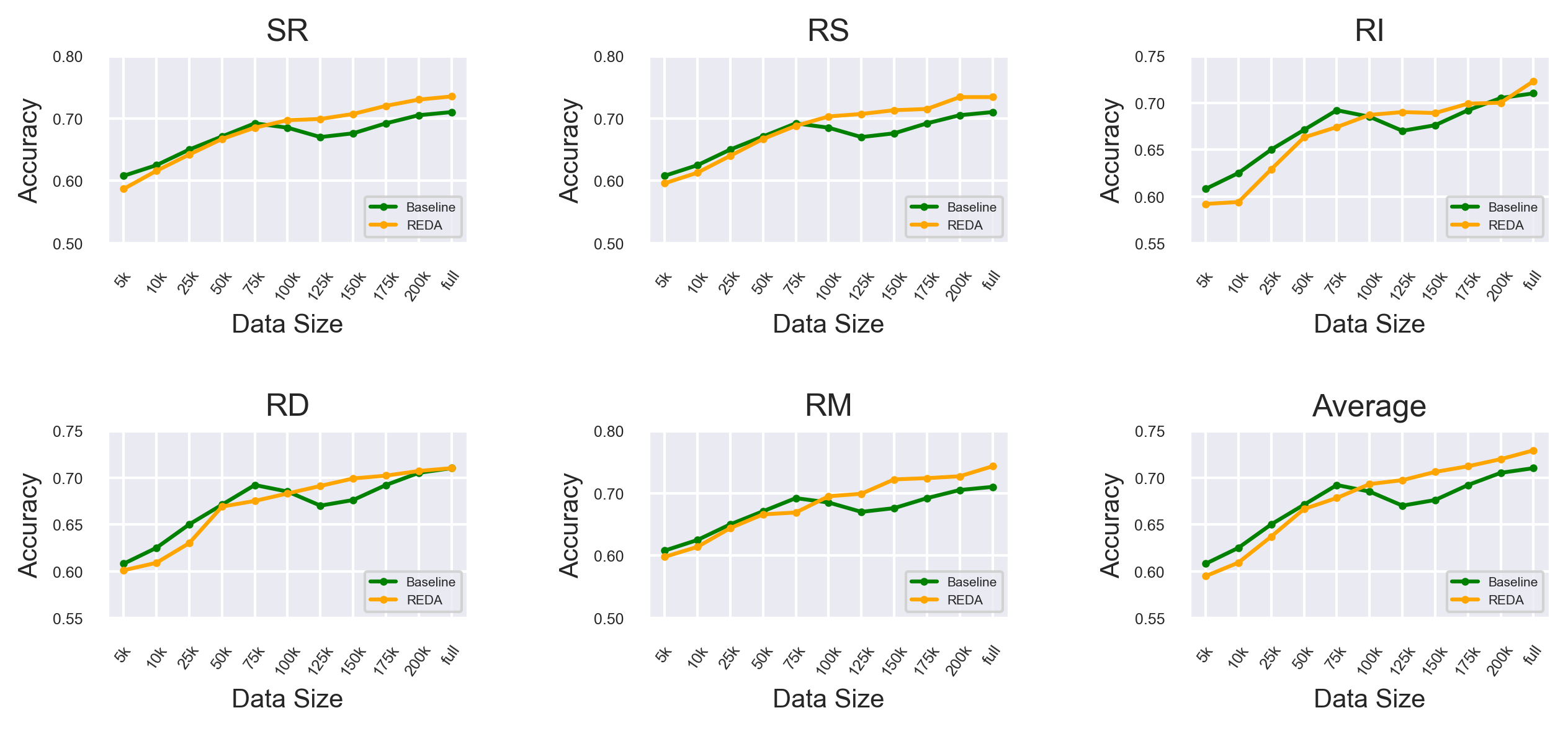}
\caption{Average test set accuracy scores of the three models under different conditions (i.e., text editing type, training data size) for the two types of LCQMC’s train sets. The sixth plot averages the statistics of the previous five plots.}
\label{fig:accuracy}
\end{figure}

Figure~\ref{fig:accuracy} shows the average accuracy scores of the three classification models trained across these 11 training sizes and under different text editing conditions. Again, it confirms that there is a threshold of training size that needs to be satisfied so that each text editing operation can boost the performance of the models. Noticeably, the threshold here appears to be the 100k training size or so, instead of 50k as in Table~\ref{tab:3}, which may have to do with the separation of these DA techniques. 

\begin{figure}[!htb]
\centering
\captionsetup{font=small}
  \includegraphics[width=1\columnwidth]{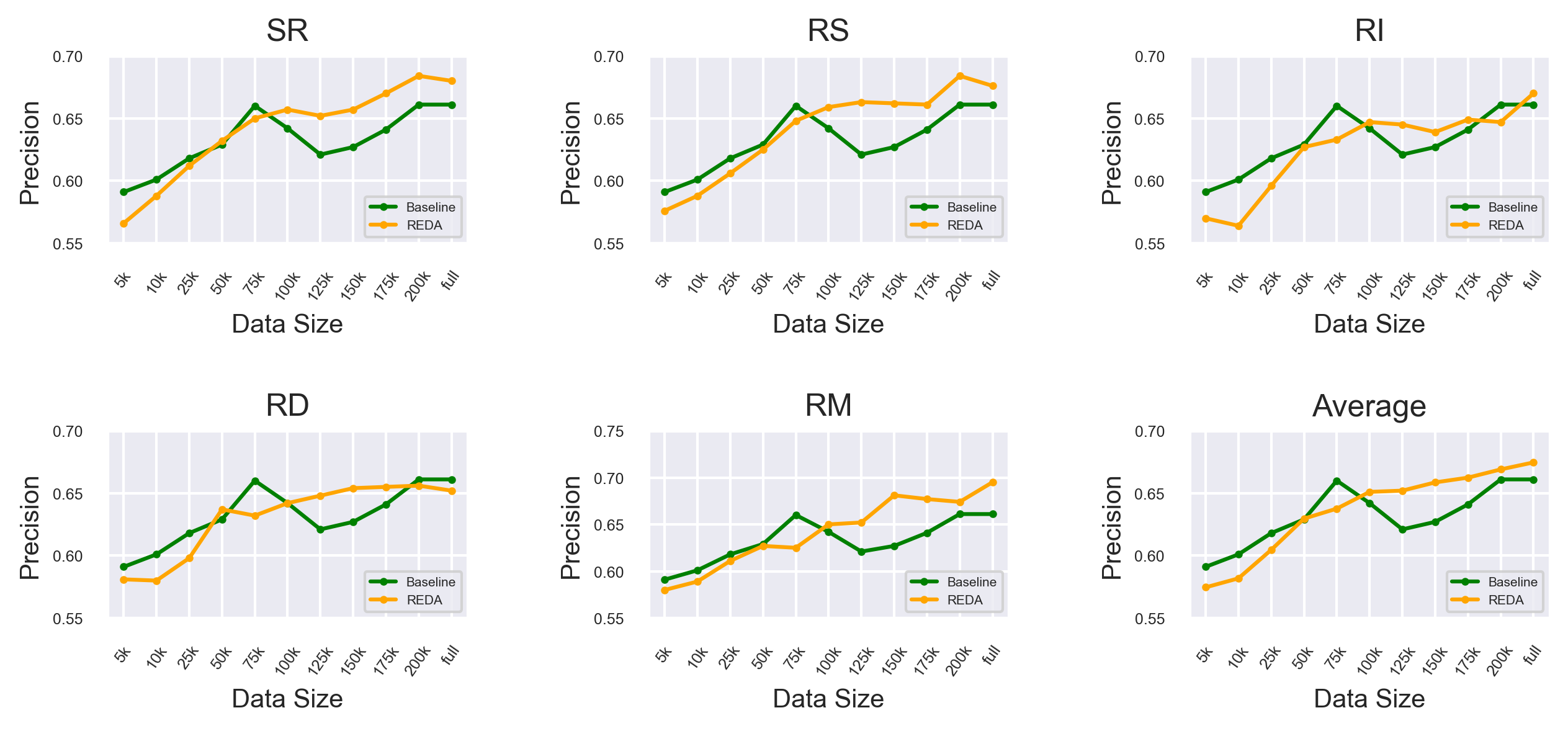}
\caption{Average test set precision scores of the three models under different conditions (i.e., text editing type, training data size) for the two types of LCQMC’s train sets. The sixth plot averages the statistics of the previous five plots.}
\label{fig:precision}
\end{figure}

\begin{figure}[!htb]
\centering
\captionsetup{font=small}
  \includegraphics[width=1\columnwidth]{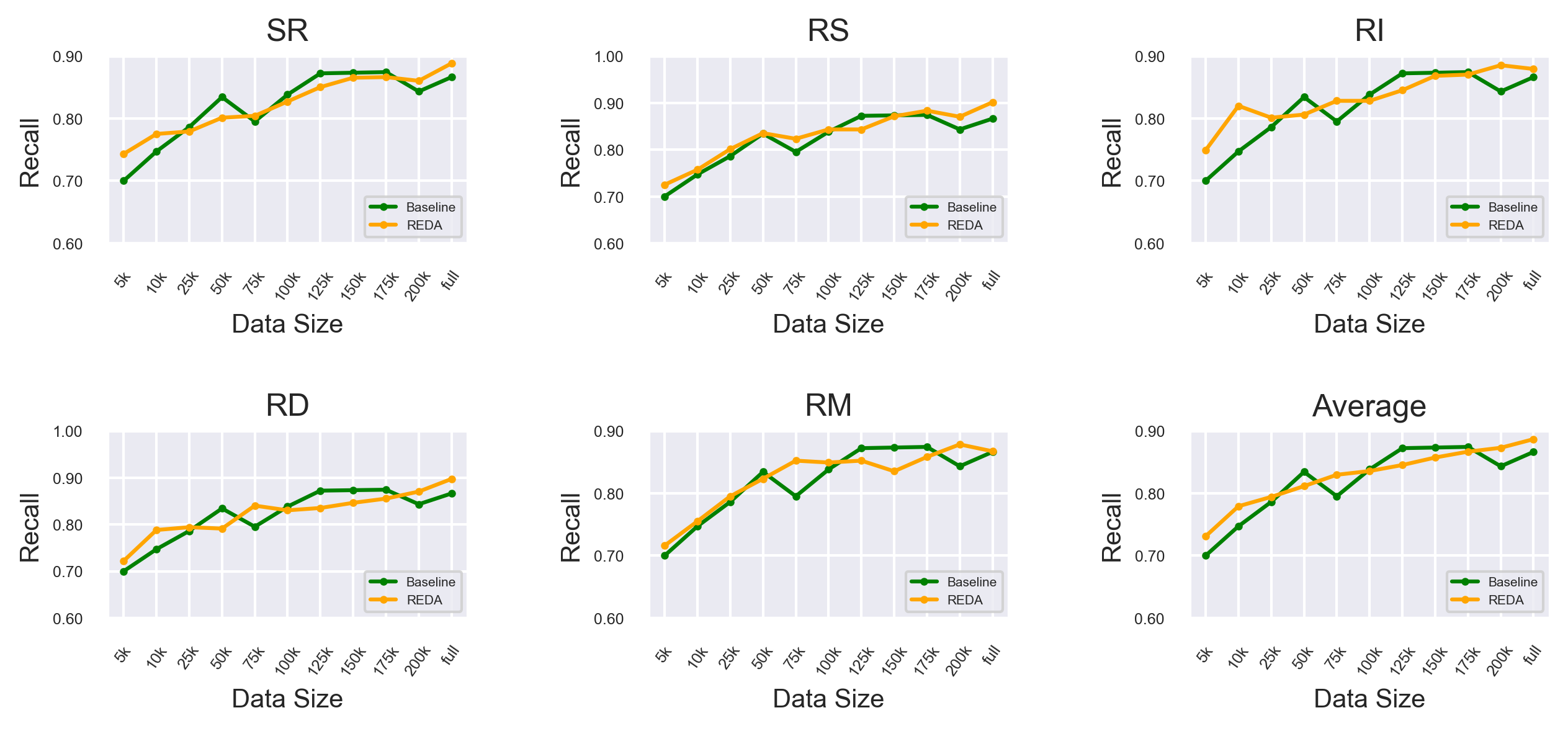}
\caption{Average test set recall scores of the three models under different conditions (i.e., text editing type, training data size) for the two types of LCQMC’s train sets. The sixth plot averages the statistics of the previous five plots.}
\label{fig:recall}
\end{figure}

To explore the possible causes for the improvement in the test set accuracy scores, we also plotted the average precision and recall scores in the same way. It turns out that the rising accuracy scores are highly correlated with the increasing precision scores, as displayed in Figure~\ref{fig:precision}, whereas such trend does not exist for the recall scores, as shown in Figure~\ref{fig:recall}, which shows more complicated patterns.

\section{\label{sec:5}Discussion and Conclusion}

In this study, we evaluate the effectiveness and generalizability of random text perturbations as a DA approach for NLP. Our experiments on binary text matching classification task in Chinese and English indicate strongly that the effectiveness of the five random text editing operations, both applied together and separately, is task specific and not generally positive. Compared to \citet{wei-zou-2019-eda} who show general success of text perturbations in simpler one-text-one-label NLP tasks across varying training sizes, we find that test set performance gains are only possible for the binary text matching task when a large amount of original training examples are seen by the models. This makes random text perturbations a less practical DA approach for text pair classification tasks, where having sufficiently large labeled data is usually expensive.  

As expected, since text matching involves classification of text pairs, the task is by nature more sensitive to the semantic changes caused by text augmentation and thus represents a more reliable way to evaluate a DA approach for NLP. The failure of random text perturbations with small train sets may imply that the classification models are misguided by the negative effects of the augmented examples, possibly related to the augmented false matching pairs, which hamper their test set performance. However, with enough original training examples supplied, the models learn to mediate these negative effects and turn them somewhat into a means of regularizations, which help the models generalize better with improving precision on the test sets. 

In relation to \citet{wei-zou-2019-eda}, another possible cause for the failure of augmentation on small train sets may have to do with the fact that REDA does not allow deduplicates to be in the augmented texts. That means, given comparably small editing rates, REDA tends to produce more diverse and yet non-paraphrastic augmented texts than EDA, which enlarges the negative effects of random text perturbations and thus demand more original training examples to mediate such effects. However, the exact theoretical reasons behind are worth further studying in the future. 

Thoroughly evaluating a DA approach for NLP is not easy. There certainly remains a lot to be done so that we can better understand and leverage the effective sides of random text perturbations, or any other DA approaches/techniques for NLP. For example, future experiments may want to examine how a model’s configurations (e.g., whether initialized with pretrained word embeddings, model architecture, hyperparameters) or the initialization of REDA may affect the test set performance for NLP tasks of various natures, e.g., classification or non-classification, binary or multi-class etc. In addition, since language is a complex discrete system, a fair evaluation also requires a large enough test set, either from one domain or across domains such that the evaluation results are more reliable and revealing. We hope this study will inspire more in-depth experiments to contribute to text augmentation, or more broadly, the empirical (evaluation) methods for NLP.

\bibliography{references}
\bibliographystyle{acl_natbib}

\clearpage

\section*{Appendix}

\subsection*{\label{sec:A}A. Initialization of REDA}

We initialized REDA with the following editing rate for SR, RS, RI, and RD, respectively: 0.2, 0.2, 0.1, and 0.1. We applied Python rounding rule to calculate and perform the number of edits needed for each operation. That means, if the number of edits is less than or equal to 0.5, it will be rounded down to 0 and thus no editing operation will apply. To make our experiments more controlled and doable, (1) we made RM only randomly perform two of the other four editing operations with one edit each; (2) and every editing operation will produce up to 2 non-duplicated augmented texts, if the train set size is less than 50k; otherwise, there will only be one augmented text instead. Every augmented text was crossed paired with the other text that was the pair to the text being augmented with the original label kept for the augmented text pair. That means, the augmented text pairs double the number of augmented texts set for each text. These settings also apply for the ablation study. 

The synonym dictionary for English comes from WordNet\footnote{\tiny\url{https://wordnet.princeton.edu}}. The synonym dictionary for Chinese comes from multiple reputable sources through web scraping\footnote{\tiny\url{https://github.com/jaaack-wang/Chinese-Synonyms}}.

\subsection*{\label{sec:B}B. Model Training}

\noindent \textbf{Training Settings}. We reused the three simple models already constructed using Baidu’s deep learning framework paddle\footnote{\tiny\url{https://github.com/PaddlePaddle/PaddleNLP/blob/develop/examples/text_matching/simnet}}. We trained all the models in Baidu Machine Learning CodeLab on its AI Studio\footnote{\tiny\url{https://aistudio.baidu.com/aistudio/index}} with Tesla V100 GPU and 32G RAM, which the author could use up to 70 hours per week.\\

\noindent \textbf{Basic Architecture}. All the models begin with an Embedding layer that outputs 128-dimensional word embeddings. Then, the word embeddings for the text pairs each go through an encoder so that the encoded embeddings for the text pairs have same output dimensions and can be concatenated along the last axis. The concatenated embeddings run through a Linear layer, a Tanh activation function, and another Linear layer that outputs two dimensional logits. The details of the encoder configurations used for the CBOW, CNN, and LSTM models can be found at the footnote.\footnote{\tiny\url{https://github.com/PaddlePaddle/PaddleNLP/blob/develop/paddlenlp/seq2vec/encoder.py}}\\

\noindent \textbf{Other}. We did not use EarlyStopping or other similar callbacks, because that might increase the experimental costs to a point that obstructs training. Also, the effect of such a callback should be trivial as most of our models overfitted within 3 epochs.

\subsection*{\label{sec:C}C. Size of augmented train sets}

Table~\ref{tab:7} and Table~\ref{tab:8} contain size of the train sets for the first two experiments on LCQMC and QQQD and the ablation experiment on LCQMC, respectively. Please note that, for simplicity, 240k is used to refer to the full size of LCQMC, which is 238,766 to be exact. Also, due to deduplication, different text editing operations may result in augmented train sets with non-trivial difference in size, as discernible in Table~\ref{tab:8}. The reason that the ratio of the augmented train sets to the corresponding original train sets in size is different is explained in Appendix \hyperref[sec:A]{A}. 

\begin{table}[h]
\footnotesize
\centering
\begin{tabular}{ccc|ccc}
\hline
\textbf{LCQMC} & \textbf{Augmented} & & &\textbf{QQQD} & \textbf{Augmented} \\ \hline
5k & 66,267 & & & 10k & 148,341 \\ \hline
10k & 132,513 & & & 50k & 543,066 \\ \hline
50k & 563,228 & & & 100k & 1,086,063 \\ \hline
100k & 929,176 & & & 150k & 1,629,178 \\ \hline
240k & 2,218,512 & & & 260k & 2,823,733 \\ \hline

\end{tabular}
\caption{\label{tab:7}
Size of augmented train sets for the first two experiments on LCQMC and QQQD.}
\end{table}

\begin{table}[h]
\footnotesize
\centering
\begin{tabular}{cccccc}
\hline
\textbf{Size} & \textbf{SR} & \textbf{RS} & \textbf{RI} & \textbf{RD} & \textbf{RM} \\ \hline
5k & 24,402 & 24,758 & 16,733 & 16,780 & 24,859 \\ \hline
10k & 48,807 & 49,575 & 33,090 & 33,208 & 49,652 \\ \hline
25k & 122,358 & 124,040 & 83,329 & 83,592 & 124,237 \\ \hline
50k & 244,577 & 248,074 & 166,839 & 167,296 & 248,539 \\ \hline
75k & 220,843 & 223,497 & 162,563 & 162,972 & 224,026 \\ \hline
100k & 294,516 & 297,987 & 216,540 & 217,012 & 298,620 \\ \hline
125k & 368,078 & 372,536 & 270,957 & 271,552 & 373,266 \\ \hline
150k & 441,643 & 446,941 & 325,027 & 325,738 & 447,838 \\ \hline
175k & 515,229 & 521,484 & 379,352 & 380,214 & 522,535 \\ \hline
200k & 588,901 & 595,977 & 433,521 & 434,469 & 597,084 \\ \hline
240k & 703,077 & 711,631 & 517,492 & 518,664 & 712,852 \\ \hline

\end{tabular}
\caption{\label{tab:8}
Size of augmented train sets for the ablation experiment on LCQMC.}
\end{table}

\end{document}